\documentclass[fleqn,10pt]{wlscirep}
\usepackage[utf8]{inputenc}
\usepackage[T1]{fontenc}
\usepackage{orcidlink}
\usepackage[separate-uncertainty=true,range-phrase=--,multi-part-units=single,range-units=single]{siunitx}
\usepackage{bm}
\title{Machine learning for cerebral blood vessels' malformations}

\author[1]{Irem Topal \orcidlink{0000-0002-4029-9954}}
\author[2]{Alexander Cherevko}
\author[2]{Yuri Bugay}
\author[3]{Maxim Shishlenin \orcidlink{0000-0001-7408-724X}}
\author[4]{Jean Barbier \orcidlink{0000-0002-2652-6727}}
\author[5,6]{Deniz Eroglu \orcidlink{0000-0001-6725-6949}}
\author[4]{\'{E}dgar Rold\'{a}n \orcidlink{0000-0001-7196-8404}}
\author[7,*]{Roman Belousov \orcidlink{0000-0002-8896-8109}}
\affil[1]{University of Padova, Department of Biomedical Sciences, Padova 35121, Italy}
\affil[2]{Lavrentyev Institute of Hydrodynamics, Novosibirsk 630090, Russia}
\affil[3]{Sobolev Institute of Mathematics, Novosibirsk 630090, Russia}
\affil[4]{ICTP---The Abdus Salam International Centre for Theoretical Physics, Trieste 34151, Italy}%, Section of Quantitative Life Sciences
\affil[5]{Kadir Has University, Faculty of Engineering and Natural Sciences, Istanbul 34083, T\"{u}rkiye}
\affil[6]{Imperial College London, Department of Mathematics, London SW7 2AZ, United Kingdom}
\affil[7]{EMBL---European Molecular Biology Laboratory, Cell Biology and Biophysics Unit, Heidelberg 69117, Germany}

\affil[*]{roman.belousov@embl.de}

\begin{abstract}
Cerebral aneurysms and arteriovenous malformations are life-threatening hemodynamic pathologies of the brain. While surgical intervention is often essential to prevent fatal outcomes, it carries significant risks both during the procedure and in the postoperative period, making the management of these conditions highly challenging. Parameters of cerebral blood flow, routinely monitored during medical interventions or with modern noninvasive high-resolution imaging methods, could potentially be utilized in machine learning-assisted protocols for risk assessment and therapeutic prognosis. To this end, we developed a linear oscillatory model of blood velocity and pressure for clinical data acquired from neurosurgical operations. Using the method of Sparse Identification of Nonlinear Dynamics (SINDy), the parameters of our model can be reconstructed online within milliseconds from a short time series of the hemodynamic variables. The identified parameter values enable automated classification of the blood-flow pathologies by means of logistic regression, achieving an accuracy of \SI{73}{\%}. Our results demonstrate the potential of this model for both diagnostic and prognostic applications, providing a robust and interpretable framework for assessing cerebral blood vessel conditions.
\end{abstract}
\begin{document}

\flushbottom
\maketitle
\thispagestyle{empty}

% \noindent Please note: Abbreviations should be introduced at the first mention in the main text – no abbreviations lists. Suggested structure of main text (not enforced) is provided below.

\section*{Introduction}

Intracranial arterial aneurysms (AA) and arteriovenous malformations (AVM) are cerebral blood-flow defects that can lead to life-threatening hemorrhage or neurological pathologies \cite{Brisman2006,AlShahi2001}. Often these threatening conditions warrant surgical intervention. However, due to the significant uncertainty of the risks associated with both, the pathologies and their treatment, determining whether such an intervention is advisable is a challenging medical question even for experienced specialists, who must judge on a wide range of complex factors.

AAs represent swollen areas caused by weakened blood-vessel walls, which can expand and rupture. No unified theory of their genesis exists, though most researchers attribute aneurysms to degenerative changes in the vascular walls caused by various factors. An aneurysm consists of a neck, body, and dome, with the dome formed by a single layer of intima (thinnest)---the most vulnerable part where ruptures typically occur. The risk of hemorrhage from an unruptured aneurysm is about \SI{1}{\%} per year, but a recurrent hemorrhage is significantly more frequent: \SIrange{15}{25}{\%} within the first two weeks and \SI{50}{\%} within the first six months. The risk of hemorrhage also increases with the size of the aneurysm: defects under \SI{5}{mm} have a lifetime rate of \SI{2.5}{\%}. Aneurysms of \SIrange{6}{10}{mm} in size burst in \SI{41}{\%} of the cases, and those in the range of \SIrange{11}{15}{mm}---in \SI{87}{\%}. Repeated hemorrhages are generally more severe than initial ones, with a fatality rate of \SI{32}{\%} in the first week, \SI{43}{\%} by the second week, and up to \SI{63}{\%} within the first year.
Surgical treatment for cerebral AAs involves either open aneurysm clipping or conservative endovascular methods, though the latter is only applicable to the pathologies of a small size.

AVMs are congenital pathologies manifested as entangled blood vessels that disrupt the oxygen supply of the surrounding tissues, thereby provoking cell death. In severe cases, they can rupture, resulting in intracranial hemorrhage with the frequency reported in the range \SIrange{30}{82}{\%}~\cite{Mast1995}. The high morbidity and disability rates highlight the dangers associated with this pathology. A common method of treating AVMs is embolization: embolizing agents, such as N-butyl-2-cyanoacrylate or a non-adhesive copolymer of ethylene and vinyl alcohol (ONYX)~\cite{Solomon2017}, are delivered by a micro-catheter to block the pathological vessels.
The mortality rate of treated patients with AVMs achieves \SI{9.3}{\%} during the period of hospitalization.

The decision on the surgical intervention requires a careful risk assessment before and after the treatment, as a re-operation may be deemed necessary~\cite{Martin1990,Thompson1998,Platz2014}. Therefore, modeling hemodynamic flows in patients with AAs and AVMs is a mathematical problem of utter practical importance.

\begin{figure}[ht]
    \centering
    \includegraphics[width=\linewidth]{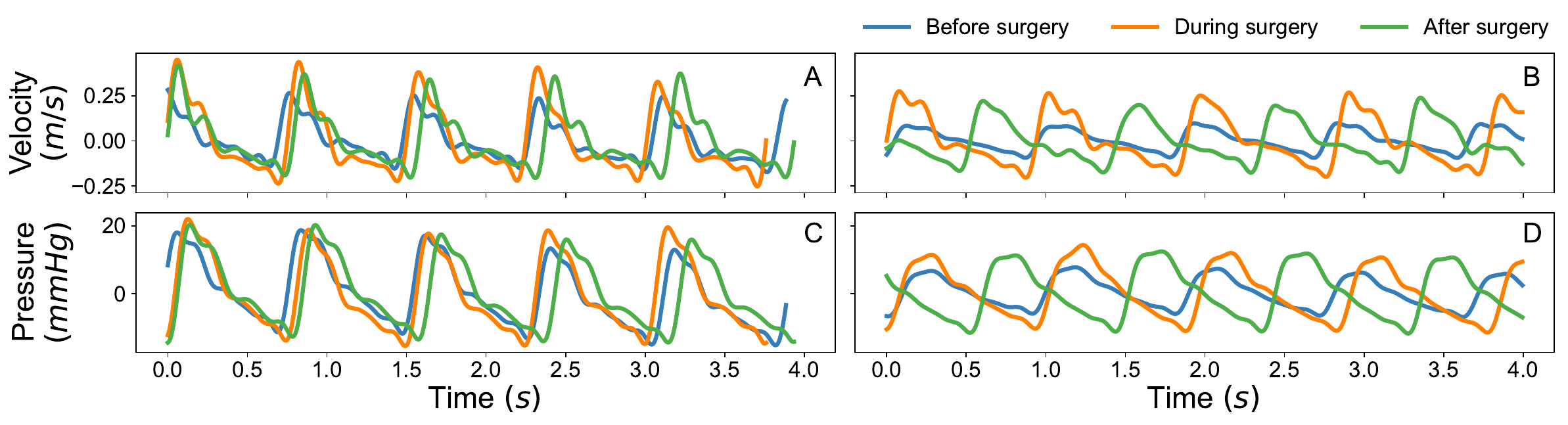}
    \caption{Examples of time series for blood flow velocity (A and B) and pressure (C and D) in clinical data before, during, and after surgery: patient with an arterial aneurysm (A and C), and patient with an arteriovenous malformation (B and D).}
    \label{fig:series}
\end{figure}

With the increasing role of computational methods in medicine~\cite{Arsalidou2022}, there is a growing interest in applying machine learning to clinical data to assist medical decision-making, treatment planning, and execution~\cite{pedrett2023technical}. In this report, we explore the use of modern machine learning techniques on data acquired during the treatment of AAs and AVMs in real-time. During surgical interventions, physical parameters of patients' blood flow, such as velocity $v(t)$ and pressure $p(t)$, are continuously monitored~\cite{orlov2014,Cherevko2016,Parshin2016}, as illustrated in Figure~\ref{fig:series}.
In previous studies~\cite{Cherevko2016,Parshin2016}, the time series of these hemodynamic variables were modeled using a nonlinear equation of the Lienard type~\cite{Strogatz2018}:
\begin{equation}\label{eq:original}
    \ddot{p}(t) + A[p(t)]\,\dot{p}(t) + B[p(t)] p(t) = \epsilon v(t),
\end{equation}
where $A(p) = a_0 + a_1 p + a_2 p^2 + \cdots$ and $B(p) =b_0 + b_1 p + b_2 p^2 + \cdots$ are algebraic polynomials in $p(t)$ with coefficients $a_{0,1,2\dots}$ and $b_{0,1,2,\dots}$.

Simulations of equation~\eqref{eq:original} represent the direct (well-posed) Cauchy problem with the initial-value conditions $p(0)=p_0$ and $\dot{p}(0)=\dot{p}_0$. In Refs.~\citeonline{Cherevko2016} and \citeonline{Parshin2016} the coefficients of algebraic polynomials $a_{0,1,2\dots}$, $b_{0,1,2,\dots}$ and the initial conditions $p_0$ and $\dot{p}_0$ are determined from experimental data $[p(t_i), v(t_i)]_{i=0,1,2,...n}$ using techniques from inverse-problem theory~\cite{Kabanikhin-2012,Kabanikhin-Shishlenin-2019}. However, this approach is computationally expensive and sensitive to initial parameter guesses~\cite{Hanke-Neubauer-Scherzer-1995,Kabanikhin-Shishlenin-2008}, as it often involves iterative, high-dimensional optimization, especially with nonlinear polynomial terms, making it costly.

Common regression techniques, including Least Squares, LASSO, and Ridge regression, often face limitations in real-time applications due to high computational demands or sensitivity to initial parameter guesses~\cite{StatLearn}. Moreover, fitting data to a minimal model risks misinterpretation, as patient-specific variations may not be adequately captured by a universal set of terms. A robust approach would, therefore, benefit from starting with a larger model framework that can adaptively reduce complexity based on individual patient data. To address the computational challenge, $L_2$-norm-based approaches, like Ridge regression, are often employed for their speed. However, while effective at controlling over-fitting, they do not inherently reduce the number of terms, resulting in models that may remain unnecessarily complex. Conversely, $L_1$-norm-based methods such as LASSO can achieve sparsity by eliminating terms with minimal contribution, but their iterative nature makes them too slow for real-time applications.

With the rapid improvement in dataset quality and methods of their processing, the data-driven discovery of dynamical systems offers an alternative approach to modeling and understanding complex real-world systems~\cite{brunton_kutz_2019}. Expanding on the previous works~\cite{orlov2014,Cherevko2016,Parshin2016}, we explore this approach and present advances in two directions: data-driven modeling of the blood-flow dynamics with real-time identification of the system parameters, and the potential clinical applications of the relationships learned from this data.

To address the limitations of the initial model, we expanded the basis functions by constructing a more comprehensive candidate library, $\bm{\Theta}$, which includes not only polynomials in $p(t)$ and its time derivatives, but also other nonlinear terms:
\begin{equation}\label{eq:general}
    \ddot{p} + c_{01} \dot{p} + c_{02} \dot{p}^2 + c_{03} \dot{p}^3
        + (c_{10} + c_{11} \dot{p} + c_{12} \dot{p}^2) p
        + (c_{20} + c_{21} \dot{p}) p^2 = \epsilon v,
\end{equation}
where $c_{ij}$ are the coefficients to be determined. This expanded equation allows us to assess the contribution of higher-order terms to the accuracy of the model and determine whether the original model is optimal in terms of fit and simplicity.

To determine the coefficients $c_{ij}$, we apply the Sparse Identification of Nonlinear Dynamical Systems (SINDy) method~\cite{Brunton2016} with Sequentially Thresholded Least Squares (STLS) optimization which is computationally efficient and converges to a sparse solution rapidly as an alternative to the traditional regression techniques. By adjusting the sparsity threshold we systematically eliminate terms, with the original equation (equation~\ref{eq:original}) being recoverable for appropriate threshold values. This trade-off between simplicity and computational feasibility makes SINDy a suitable method for real-time applications.  A detailed explanation of the SINDy method is provided in Methods.

In the regression procedure, we simplify the inverse problem by setting $p(0)$ to the experimental value $p_0 = p(t_0)$, estimating $\dot{p}(0)$ from the forward-difference formula $p_0 = [p(t_1) - p(t_0)] / (t_1 - t_0)$, thus excluding these values from the fitting parameters. By setting the sparsity threshold at $\eta = 1.0$, the nonlinear terms of order $O(p^3)$ are eliminated, and by $\eta = 5.0$, equation~\eqref{eq:general} is reduced to a linear model of a forced damped harmonic oscillator
\begin{equation}\label{eq:model}
    \ddot{p}(t) + a \,\dot{p}(t) + b \,p(t) = \epsilon v(t),
\end{equation}
where $a$, $b$ and $\epsilon$ are constants. This linear approximation retained only three key parameters---$a$, $b$, and $\epsilon$---resulting in a model well suited for real-time clinical application due to its balance of simplicity and accuracy.

When revisiting the original polynomial formulation in equation~\eqref{eq:original}, we observed that the constant terms $a_0$ and $b_0$ dominated in the trend of $A(p)$ and $B(p) p$ averaged over patients, while higher-order terms were negligible (Supplementary Figure~\ref{fig:si:polynomials}), supporting the suitability of the linear model in equation~\eqref{eq:model} to capture the core dynamics of blood flow.

Through SINDy, the computational expense of handling general models becomes manageable, and sufficiently increasing the sparsity threshold further allows for simplification without losing fitting accuracy and reflects a successful complexity reduction. This simplified model enhances robustness to noise in clinical data, making it well-suited for real-time applications.

In summary, our findings indicate that the complex nonlinear model proposed in Refs.~\citeonline{Cherevko2016} and \citeonline{Parshin2016} may be overly detailed. The simplified linear model, with only three parameters, effectively captures the essential dynamics of blood flow and offers a more practical, interpretable solution. We specifically evaluated the reproducibility of parameter values across datasets, observing that, even in the worst case, the relative deviation of the parameters did not exceed \SI{24}{\%} (see Results). In addition, by reducing the fitting problem to solving an overdetermined linear system, SINDy does not require an initial parameter guess, which is one of the critical aspects in the original technique of Refs.~\citeonline{Cherevko2016} and \citeonline{Parshin2016}.

Further exploration is needed to assess whether the fitted values of the model parameters, viewed as means of data-dimensionality reduction, could act as predictor variables for pathologies or treatment prognosis in higher-tier machine learning models. To pursue this line of inquiry, we applied logistic regression to classify blood-flow anomalies and surgically treated vessels based on the parameters of equation~\eqref{eq:model} inferred from the patient data. In this application, our automatic classifier discriminates between three classes---flows with AA, AVM, and in surgically treated vessels---with an accuracy of 73-\si{\%}, which is a surprisingly promising result given the limited data.

The proposed model offers a novel approach to analyzing cerebral blood flow in patients undergoing surgery for cerebral blood vessel malformations. Using the inferred parameters of the model in surgical planning could help medical doctors make more informed decisions. Additionally, our findings highlight the prospects of using larger datasets to design automatic diagnostic tools for blood pathologies or methods for forecasting the outcome of a surgical treatment. Although our study relies on direct measurements, which have the advantage of being more precise, noninvasive techniques of measuring blood-flow variables also exist~\cite{vanZijl1998,Lee2004,Schuchardt2015,Errico2015,Morgan2020} and could be used for therapeutic assessment before the intervention stage.

\section*{Materials and Methods}

In this study, we employ a multi-step approach to analyze clinical data and apply machine learning for classification. First, the clinical data is reconstructed using sparse regression methods, which help identify relevant features while reducing model complexity. Next, we apply logistic regression to classify the reconstructed data into distinct categories of blood flow anomalies, based on clinical outcomes. The following subsections describe the details of the \textit{data acquisition}, the \textit{sparse regression procedure}, and the \textit{classification methodology}.

\subsection*{Clinical data}

The clinical data used in this report were acquired during neurosurgical operations at the Meshalkin Research Institute of Circulation Pathology (Novosibirsk, Russia). Blood velocity and pressure were recorded using a Doppler sonography from an intravascular guidewire with a diameter \SI{0.34}{mm}. The signals were processed through an analogue-to-digital converter at a frequency of 200-\si{Hz}. A noise filter was applied to eliminate Fourier components, whose frequencies exceed the nominal time resolution of the device. We collected velocity and pressure data before, during, and after surgery from ten patients: five with AAs, and five with AVMs, all of whom underwent a successful surgical treatment.

Each time series lasted between \SIrange{3.345}{5.375}{s}, with a time step of $\Delta t = \SI{5}{ms}$, corresponding to approximately five cardiac cycles per patient. The time-averaged values of pressure and velocity were subtracted from each trajectory before fitting the data to theoretical models.

\subsection*{Sparse identification of blood-flow dynamics}

Many dynamical systems of variables $\bm{x}(t) \in \mathbb{R}^n$ are described by governing equations of the form $\dot{\bm x}(t) = \bm{f}[\bm{x}(t)]$, which contain just a few terms in the expression for $\bm{f}(\bm{x})$. Therefore, such equations are sparse in some basis (set) of functions. SINDy was motivated by this observation and aims to solve this sparse identification problem~\cite{Brunton2016, desilva2020pysindy}. In the same spirit, given an observed dependency of velocity on time, we are looking for a governing equation of pressure dynamics in the form of a second-order differential equation:
\begin{equation}
    \ddot{p}(t) = {f}[p(t), \dot{{p}}(t), v(t)],
    \label{second-order}
\end{equation}
given the initial values $p(0)$ and $\dot{p}(0)$.

In vector form, we denote the time series as $\bm{P} = \left[p(t_0), p(t_1),\dots, p(t_n)\right]$, $\bm{\dot P} = \left[\dot p(t_0), \dot p(t_1),\dots\right]$, and $ \bm{V} = \left[v(t_0), v(t_1),\dots\right]$ for the consecutive instances of time $t_{i=0,1,2,\dots,n}$.
By performing a sparse regression on the linear equation 
\begin{equation}
    \ddot{\bm {P}} = \bm \Theta (\dot {\bm P}, \bm P, \bm V)\, \bm \Xi,
\label{eq:linear_eq}
\end{equation}
we seek a vector of coefficients, $\bm \Xi$, that identifies the system's dynamics within the basis of candidate functions such as $p(t)$, $\dot{p}(t), p^2(t)$, and others. These candidate functions are organized into a library, represented by the matrix
\begin{equation}
 \bm{\Theta}(\dot {\bm P}, \bm P, \bm V) = \begin{bmatrix}
\vrule & \vrule & \vrule & \vrule & \vrule & &\vrule\\
 {p}(t) & {\dot p}(t) & {p}(t) {\dot p}(t) & {p}(t)^2 & {\dot p}(t)^2& \dots & {v}(t)  \\
 \vrule & \vrule & \vrule & \vrule  & \vrule & & \vrule \end{bmatrix}\in\mathbb{R}^{n+1\times k},
\end{equation}
with each column containing one of the $k$ basis functions evaluated at the consecutive instances of time $t_{i=0,1,2,...,n}$. To compose $\bm{\Theta}$ we approximate the derivative of the pressure numerically $\dot{p}(t)$ as
\begin{equation}
    \dot{p}(t_i) = \begin{cases}
        \frac{p(t_{i+1}) - p(t_{i-1})}{t_{i+1} - t_{i-1}} + O(\Delta{t}^3)\text{, if $0 < i < n$}\\
        \frac{p(t_1) - p(t_0)}{t_1 - t_0} + O(\Delta{t}^2)\text{, for $i=0$},\\
        \frac{p(t_n) - p(t_{n-1})}{t_n - t_{n-1}} + O(\Delta{t}^2)\text{, for $i=n$},
    \end{cases}
\end{equation}
and similarly to calculate $\ddot{\bm P}$ given this approximation of $\dot{\bm P}$. The central finite differences in the above equation are accurate up to the second order in $\Delta{t} = t_{i+1} - t_{i}$ for all instants $t_{0 < i < n}$, except the boundaries $t_0$ and $t_n$, where it is not applicable, and we have to resort to the first-order forward and backward finite differences, respectively. The coefficients $\bm \Xi$ are determined by STLS, which starts with an ordinary least squares solution
\begin{equation}
{\hat{\bm{\Xi}}} = \underset{{\bm \Xi}}{\arg\min} \left(\|\ddot{\bm P} - \bm \Theta (\bm P, \bm V) {\bm \Xi} \|_2 \right),
\end{equation}
and eliminates the coefficients $\bm \Xi$ that are smaller than the pre-defined sparsity threshold $\eta$. Then, another least squares solution is obtained for $\bm \Xi$ on the remaining coefficients, which are compared to the threshold again. This procedure is iterated until no more coefficients can be zeroed out.

By increasing the sparsity threshold $\eta$, we can select progressively simpler models with fewer significant non-zero terms. In our case, larger thresholds lead to models that consistently highlight the same key terms across different patients, as their data share similar patterns. This ensures that the model captures the most universally relevant dynamics without over-fitting~\cite{Mehta2019}. We have tested values of $\eta \in \{0.1,\,1.0,\,5.0\}$.

The accuracy of identified models can be characterized by the root mean squared error (RMSE): 
\begin{equation}\label{eq:mse}
    {\rm RMSE}(p, \hat p) = \sqrt{\frac{1}{n + 1} \sum_{i=0}^{n} \left[p(t_i) - \hat{p}(t_i)\right]^2}.
\end{equation}
where $p(t)$ represents the ground-truth time series of pressure, while $\hat{p}(t)$ is the trajectory generated by the identified model, starting from the same initial conditions, $\hat{p}(0) = p(0)$ and $\dot{\hat{p}}(0) = \dot{p}(0)$, and using the best-fit parameter values,  $\hat{\bm{\Xi}}$. This RMSE measure provides a way to assess how closely the identified model replicates the observed data over time.

To benchmark the computational cost of fitting, we measured the time taken for seven runs of the optimization routine, each consisting of $10^3$ iterations, executed on a single core of an Intel Core i7-13700 processor.

\subsection*{Logistic Regression for Classification of Blood-Flow Dynamics}

Using several hundred elements of two-dimensional time series of the velocity and pressure $\left(p(t_i), v(t_i)\right)$, we identify between three and ten coefficients $\hat{\bm{\Xi}}$ depending on the assumed model, effectively reducing the number of \textit{descriptors} representing the same data.

However, features of blood flow differ between the experimental conditions, i.e. before, during, or after the surgery (Figure~\ref{fig:series}). Since the pressure-velocity relationship is encoded in the parameters $\hat{\bm{\Xi}}$, these learned parameters from the patients' data can be used as inputs in a classification problem. Here, we consider three categories of interest for each patient
\begin{equation}
\label{eq:classification}
    Y(\hat{\bm{\Xi}}^{(i)}) = \begin{cases}
    0 \quad \text{for AA flow}, \\
    1 \quad \text{for AVM flow}, \\
    2 \quad \text{for flow after a successful treatment},
    \end{cases}
\end{equation}
where $Y(\hat{\bm{\Xi}}^{(i)})$ refers to the actual class of the patient $i$ and the last category is interpreted as a pathology-free condition.

In the context of a multi-class logistic regression~\cite{StatLearn,Mehta2019}, which we adopt in this report, the probability for the parameter values $\hat{\bm{\Xi}}^{(i)}$ of $i$\textsuperscript{th} patient to belong to the category $m \in \{0,1,2\}$ can be encoded by the softmax function as
\begin{equation}\label{eq:softmax}
P_m(\hat{\bm{\Xi}}^{(i)}) = \frac{\exp\left(
        \sum_j w_{mj} \hat{\Xi}_j
    \right)}{\sum_{m=0}^2 \exp\left(\sum_j w_{mj} \hat{\Xi}_j\right)}
\end{equation}
where the summation over $j$ includes all input coefficients $\hat{\Xi}_j$, namely $a$, $b$ and $\epsilon$. The input arguments $\bm{Xi}_j$ are extended by a bias term $\hat{\Xi}_0 = 1$, and $w_{mj}$ are the weights that the model optimizes to assign categories. To determine these weights, maximizing the multinomial log-likelihood function~\cite{Mehta2019}
\begin{equation}\label{eq:loglike}
    \hat{\bm{w}} = 
    \underset{\bm{w}}{\arg\max}\sum_i\sum_{m=0}^2 \Bigg\{
        y^{(i)}_{m} \ln P_m(\hat{\bm{\Xi}}^{(i)}) + (1 - y^{(i)}_m) \ln \left[1-P_m(\hat{\bm{\Xi}}^{(i)})\right]
    \Bigg\},
\end{equation}
we use the L-BFGS algorithm~\cite{lbfgs}, which ensures efficient and reliable convergence to the optimal values $w_{mj}$. In equation~\eqref{eq:loglike} we set $y^{(i)}_{m} = 1$ if $i$\textsuperscript{th} time series has the label $m$, $Y(\hat{\bm{\Xi}}^{(i)}) = m$, $y^{(i)}_{m} = 0$ otherwise.

\section*{Results}

\subsection*{Efficient identification of blood-flow dynamics}

To identify the most efficient model from the same family as equation~\eqref{eq:original}, we progressively increased the sparsity threshold $\eta$. This process starts with a more general candidate library $\bm{\Theta}$---equation~\eqref{eq:general}, including polynomials of the pressure and its time derivatives up to the third degree, along with a linear term proportional to the velocity of the blood flow.  We varied the sparsity threshold $\eta$ over a range of values, from a small threshold $\eta = 0.1$ and an intermediate value $\eta = 1.0$ to the maximum $\eta = 5.0$.

The threshold value $\eta=0.1$ narrows down the candidate library to the form of equation~\eqref{eq:original} for all patients' data (Supplementary Figure~\ref{fig:si:fitting}A). The majority of the nonlinear terms disappear at $\eta = 1.0$ (Supplementary Figure~\ref{fig:si:fitting}B). Finally, the linear equation~\eqref{eq:model} emerges robustly across all patients' data at the threshold $\eta = 5.0$ (Supplementary Figure~\ref{fig:si:fitting}C).

\begin{figure}[ht]
    \centering
    \includegraphics[width=\linewidth]{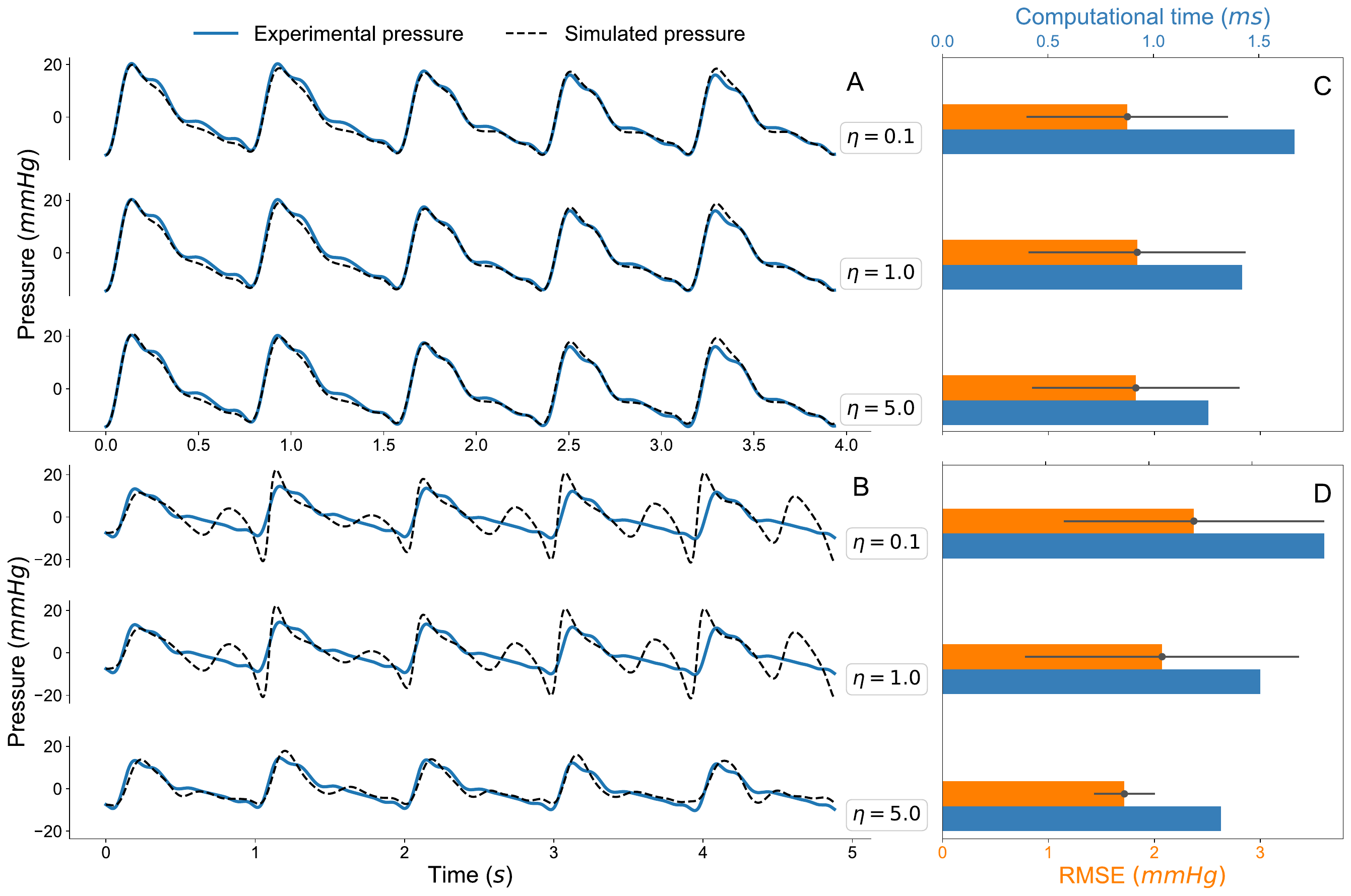}
    \caption{Illustration of the fitting procedure. With larger sparsity threshold $\eta$ the results converge to sparser models. Simulated pressure trajectories from these learned models are compared with the experimental observations. As illustrative examples we select the patients, whose models show the worst match with the experiments by fitting accuracy across all AA (A) and AVM (B) patients. The bars aligned to the threshold values summarize the overall performance of the learned model quantified by RMSE, averaged over all patients with the same pathology (orange bars, standard deviation shown by whiskers), and computational time (blue bars, average over several runs for the patients in A and B). Higher-order models show average RMSE similar to the simplest linear equation~\eqref{eq:model} for AA patients (C). However, on AVM patients (D) the linear model performs strikingly better than the alternatives. For both AA and AVM data, the linear model prevails by the computational cost (blue bars in C and D)}
    \label{fig:library}
\end{figure}

To validate the identified models, the system was simulated using the inferred parameters. The time series thus generated were then compared against the experiments (Figure~\ref{fig:library}). While the sparsest solution demonstrates the best performance, reinforcing its utility for generalization, on certain patients' data the higher-order models may appear indistinguishable from it (Figure~\ref{fig:library}A), or even strikingly inefficient (Figure~\ref{fig:library}B).

The inefficiency of higher-order equations may seem counterintuitive, as one might expect more complex models to perform better at intermediate thresholds. However, it can be explained by the fitting process, which aims to reproduce the pressure derivatives evaluated numerically rather than the original time series. The higher-order equations robustly fit the derivative data tighter than the linear model (Supplemental Figure~\ref{fig:si:derivatives} with the same patients as in Figure~\ref{fig:library}).

As the time derivatives of pressure in our method are calculated numerically from real-time series, artifacts due to noise, filtering, or smoothing in the data can occur. While advanced techniques like Kalman filtering or other smoothing methods \cite{kaptanoglu2022pysindy} could potentially reduce such artifacts and improve consistency, finding the optimal method for numerical differentiation is beyond the scope of this study. Moreover, such approaches do not always guarantee a reliable performance, particularly given the variability in clinical data. Instead, selecting a large sparsity threshold $\eta$ effectively isolates the most significant terms and ensures robustness even in the presence of artifacts related to the numerical differentiation. These features make the combination of a large threshold $\eta$ with the finite-difference technique for computing time derivatives sufficient to achieve a balance between simplicity, consistency, and accuracy.

When evaluating the statistics and performance across all our clinical data, the \emph{sparsest} model~\eqref{eq:model}, which is computationally most efficient, achieves accuracy close to the higher-order models used in the earlier works. Overall, the linear model performs strikingly better than its more complex alternatives on the AVM data as quantified by the average RMSE (Figure~\ref{fig:library}D, orange bars).

\subsection*{Reproducibility and forecasting performance of the linear model}

\begin{figure}[ht]
    \centering
    \includegraphics[width=\linewidth]{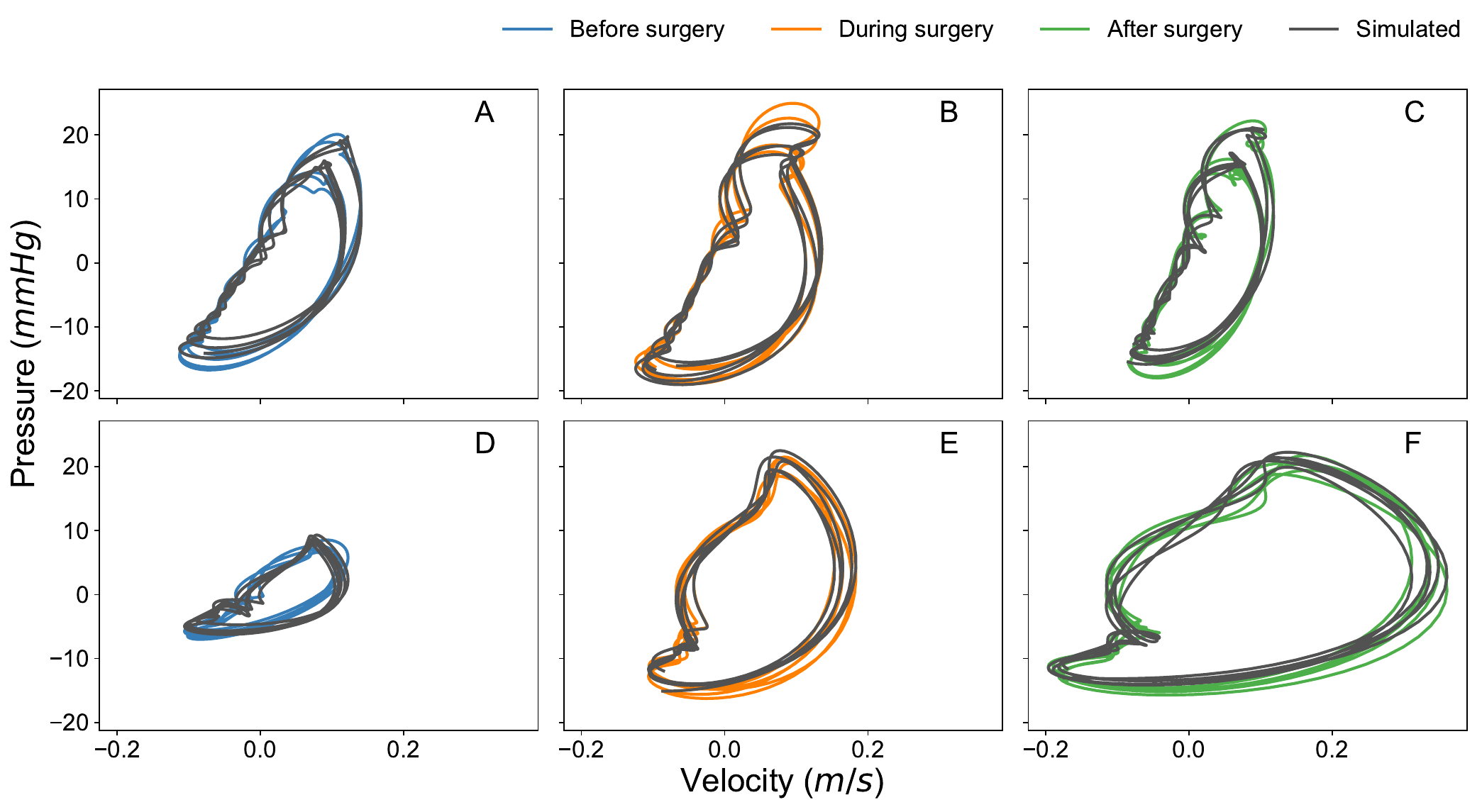}
    \caption{Examples of pressure-velocity diagrams before (blue), during (orange) and after (green) surgery for time series of a patient with an arterial aneurysm (A--C) and with an arteriovenous malformation (D--F). We simulate equation~\eqref{eq:model} for each patient by using the initial condition of the experimental pressure. The black lines show the simulated trajectory, which is in good agreement with the experimental time series, and display a counterclockwise current---the circulation commonly found in arteries~\cite{Cherevko2016,Parshin2016}.}
    \label{fig:AA-AVM_results}
\end{figure}

\begin{figure}[hb]
    \centering
    \includegraphics[width=0.9\textwidth]{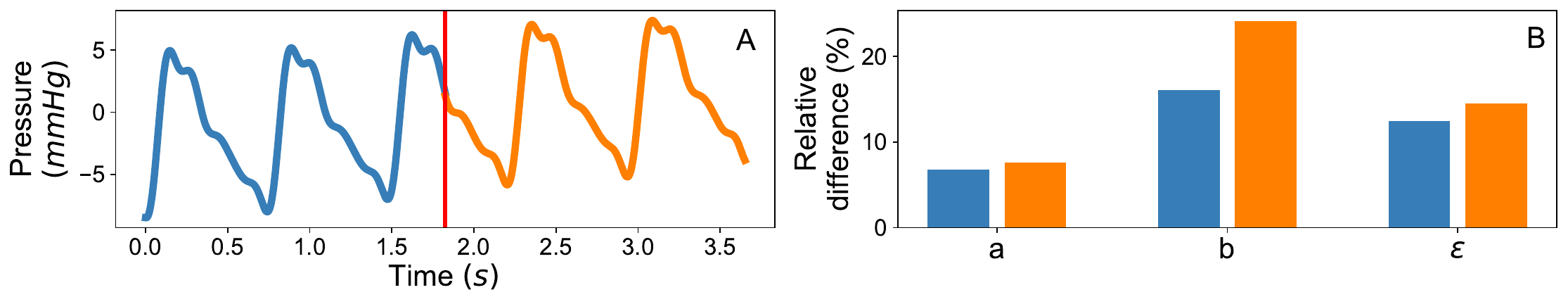}
    \caption{Reproducibility of AA patients' parameter values. A: example of splitting in half the pressure trajectory taken from an AA patient before surgery. First, we infer parameters from each part (blue and orange trajectories) independently, and then from the whole time series. B: the maximum relative difference of model parameters learned from the two halves shown in A (blue and orange) with respect to the values inferred from the entire series emphasizes the reproducibility of our results.}
    \label{fig:AA_coeff}
\end{figure}

The linear model~\eqref{eq:model} inferred by SINDy accurately captures patients' blood-flow pressure from the velocity time series (Figure~\ref{fig:AA-AVM_results}) with a set of just three coefficients $(a, b, \epsilon)$. Assessing the reproducibility of these parameters for the same patient is crucial for validating the analysis and demonstrating the robustness of the methodology. To evaluate reproducibility, we divided the time series into two equal parts and independently applied our fitting procedure to each half of the data  (Figure~\ref{fig:AA_coeff}A). Values of the parameters thus learned differed at the most by \SI{24}{\%} relative to the estimate inferred from the whole series across all patients' data considered, underscoring the robustness of the model across all patient data considered (Figure~\ref{fig:AA_coeff}B).

To test the effect of the training-set length on the forecasting performance of our method, we inferred the model parameter values from time series of varied length: we fitted a total of 1, 2, or 3 consecutive cardiac cycles, while reserving the last fifth cycle available in our data(\SI{20}{\%} of the full duration) as a test set. Thereby the fourth cycle separating the training and test sets mitigates time-correlation effects in validation. The forecasting performance is then characterized by the RMSE~\eqref{eq:mse} of the simulations reproducing the test set.

\begin{figure}[ht]
    \centering
    \includegraphics[width=\linewidth]{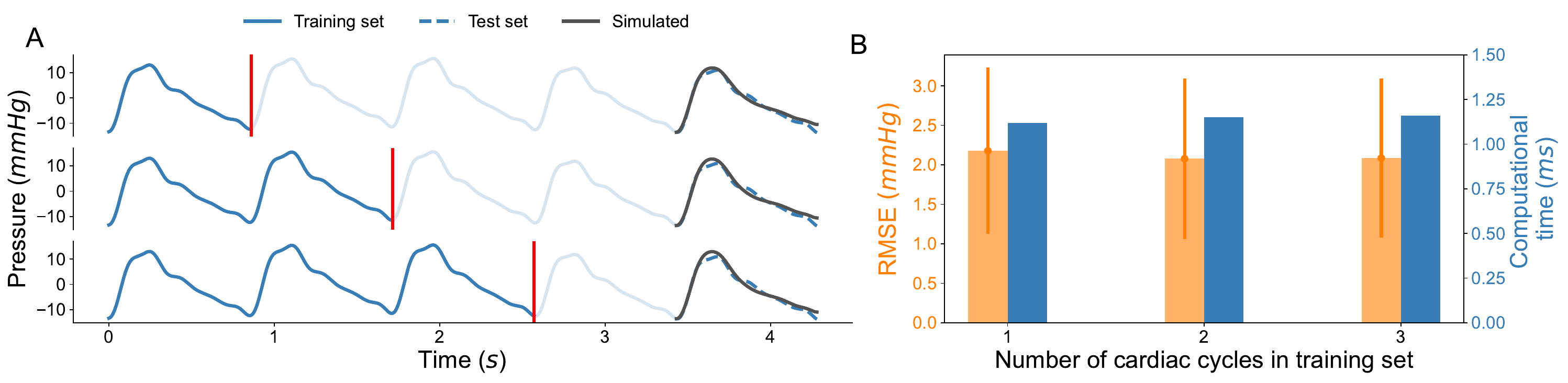}
    \caption{Forecasting performance of equation~\eqref{eq:model}. We split the pressure time series into training and test sets. Simulations of the model with parameter values learned from the training sets of various lengths are compared with the time series of the test set. A: an example of splitting an AA patient's pressure time series. The training sets (solid blue line) are delimited by the red vertical lines. The last cardiac cycle (dashed blue line), kept as the test set, is compared with the simulated model (solid black line). Averaged RMSE of the simulated pressure on a test set of a single cardiac cycle across all AA patients indicate insignificant changes in the accuracy with the increasing size of the training set (B), which expectedly offsets the computational cost (C). The orange whiskers in B indicate the standard deviation over all AA patients.}
    \label{fig:AA_tss}
\end{figure}

Training the model on a dataset covering a single cardiac cycle enables excellent predictions of the experimental time series for the same duration  (Figure~\ref{fig:AA_tss}). Extending the training set to encompass up to four cardiac cycles does not significantly affect the RMSE, though it expectedly increases the computational cost. This vindicates the SINDy method and its robustness in the present medical application due to the low variance of the experimental time series over different cycles and, thus, to low overfitting, even when training on such short time series.

\subsection*{Machine learning classification of arteriovenous pathologies}

Since the patients' data used in this work contain successfully treated blood-flow defects, the post-surgery time series of pressure and velocity may serve as proxies of healthy vessels. Consequently, we investigate whether the parameters of equation~\eqref{eq:model} learned from observations (Supplementary Tables~\ref{tab:1} and \ref{tab:2}) can be used to discriminate between the normal and abnormal flow dynamics, and to identify the pathological defect in the latter case.

Using multi-class logistic regression~\cite{StatLearn,Mehta2019}, we train the softmax classifier equation~\eqref{eq:softmax} with the feature set $\bm\Xi= (\Xi_0,\Xi_1,\Xi_2,\Xi_3) = (1, a, b, \epsilon)$, where $\Xi_0$ and its associated weight $w_{m0}$ correspond to the constant bias term (See Materials and Methods for details). The classifier is trained to categorize the output into three categories, $m \in \{0,1,2\}$: arterial aneurysm, aretriovenous malformations, and treated vessels respectively, cf. equation~\eqref{eq:classification}.

To exploit a small set of twenty examples---comprising five arterial aneurysms and five arteriovenous malformations, both pre- and post- surgery---we resample the data into 100 partitions, each with training and test subsets of 16 (\SI{80}{\%}) and 4 (\SI{20}{\%}) examples, respectively. Note that in total, there are over 4800 such unique partitions.

\begin{figure}[ht]
    \centering
    \includegraphics[width=\columnwidth]{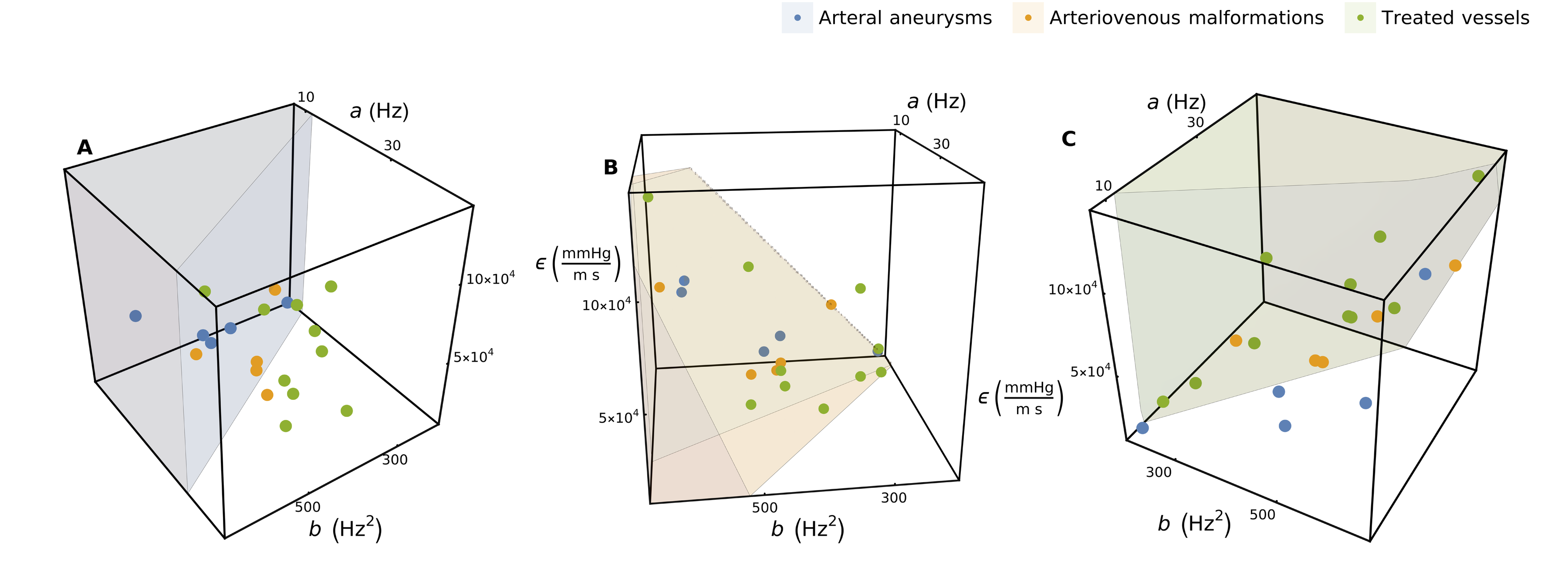}
    \caption{Decision boundaries of the logistic classifier equation~\eqref{eq:loglike} in the $3$-dimensional feature space of parameters $a$, $b$, and $\epsilon$ from equation~\eqref{eq:model}. These boundaries separate the regions of $5$ AA patients (blue circles, blue area in A) and $5$ AVM patients (orange circles, orange area in B) before the surgery from the same patients' data after the surgery (green circles, green area in C). The decision boundaries between these three classes are constructed by using Wolfram Mathematica~\cite{Mathematica}.}
    \label{fig:classification}
\end{figure}

The average accuracy of the logistic regression, evaluated over the $100$ random partitions, is \SI{73\pm2}{\%}. To investigate further properties of the classification we constructed decision boundaries by optimizing equation~\eqref{eq:softmax} over the full set of 20 examples in  in the $3$-dimensional feature space~(Figure~\ref{fig:classification}), as well as their $2$-dimensional projections (Supplementary Figure~\ref{fig:si:classification2D}).

Most of the patients' data, with a single exception, fall into the regime of an underdamped harmonic oscillator with $a^2 < b$ (Supplementary Figure~\ref{fig:si:damping}). We notice that AA occupies the region of small frequencies $\sqrt{b}$ and low dissipation $a$ (Figure~\ref{fig:classification}A). In contrast, examples of the treated vessels reside in the high-dissipation and moderate-frequencies area (Figure~\ref{fig:classification}C). Arteriovenous malformations belong in a layer of relatively smaller volume in between these two categories. In addition, boundaries of the arterial aneurysms' region appear parallel to the direction of the axis $\epsilon$ and, therefore, independent of the coupling strength between the pressure and velocity.

\section*{Discussion}

Our work illustrates the power of real-time monitoring of cardiovascular surgeries enhanced by online machine learning analysis of hemodynamic parameters. One of the main fruits of our work is the development of a fast and robust model-building procedure, which succinctly encodes the observations with just a few interpretable parameters. The combination of SINDy and a linear dynamical model renders a reliable representation of the clinical data, overcoming the computational challenges faced by previous techniques and offering excellent forecasting capabilities. Moreover, our approach is computationally efficient, making it well-suited for real-time applications, and the derived parameters allow for easy interpretation, which is crucial for various settings, including the clinical one.

The hemodynamic descriptors obtained through our procedure prove to be sufficiently informative for automated detection of pathological and normal blood flows. By using the logistic regression over a small set of training data, we achieved a~73-\si{\%} accuracy of discrimination between arterial aneurysms, arteriovenous malformations, and treated vessels. Moreover, this method, in combination with a linear model, renders it accessible to analysis and physically interpretable decision boundaries. Incorporating the model parameters into surgical planning can enhance medical decision-making and reduce the risk of unnecessary reoperation.

Our results could also inspire further efforts in the acquisition of larger patients' datasets. Beyond diagnostic applications, illustrated here by the classification of pathologies, one may pursue a more ambitious goal of predicting how the hemodynamic descriptors of an abnormal blood flow change after surgical intervention. Such predictions would facilitate the development of a therapeutic prognosis, potentially aiding medical decisions and improving patient outcomes.

\bibliography{refs}

% \section*{Data availability}
% Data is deposited as supplementary information files or can be requested by email from the corresponding author.

\section*{Ethics declarations}
Measurements of blood flow velocity and pressure were performed intraoperatively. The investigation was conducted in accordance with the Helsinki Declaration and approved by the Inspection Commission of the E.N. Meshalkin National Medical Research Center (Session No. 549). All included patients gave their informed consent.

\section*{Acknowledgements}
% Acknowledgements should be brief, and should not include thanks to anonymous referees and editors, or effusive comments. Grant or contribution numbers may be acknowledged.
A.Cherevko and Yu. Bugay acknowledge the support of Russian Science Foundation (grant no. 22-11-00264).
M. Shishelenin acknowledges the support of the state assignment of IM SB RAS (Theme No. FWNF-2024-0001). 
\'E Rold\'an acknowledges financial support from PNRR MUR project PE0000023-NQSTI.
J. Barbier was funded by the European Union (ERC, CHORAL, project number 101039794). Views and opinions expressed are however those of the author only and do not necessarily reflect those of the European Union or the European Research Council. Neither the European Union nor the granting authority can be held responsible for them.
D. Eroglu was partially supported by  TUBITAK (118C236),  UKRI (EP/Z002656/1), and the BAGEP Award of the Science Academy.
R. Belousov acknowledges funding from the EMBL.
I.T., J.B., \'{E}.R., and R.B. thank the London Mathematical Laboratory school for establishing the initial framework of this collaboration.

\section*{Author contributions statement}
% Must include all authors, identified by initials, for example:
% A.A. conceived the experiment(s),  A.A. and B.A. conducted the experiment(s), C.A. and D.A. analysed the results.  
I.T. carried out the data analysis. A.C., Yu.B., M.Sh. prepared and processed the original data. J.B. and D.E. supervised the data analysis. The original project was proposed by J.B. and É.R., and coordinated by R.B. 
All authors contributed to planning the project, as well as writing and reviewing the manuscript.

\section*{Additional information}
% To include, in this order: \textbf{Accession codes} (where applicable); \textbf{Competing interests} (mandatory statement).
% The corresponding author is responsible for submitting a \href{http://www.nature.com/srep/policies/index.html#competing}{competing interests statement} on behalf of all authors of the paper. This statement must be included in the submitted article file.
The authors declare no competing interests.

\clearpage
\setcounter{figure}{0}
\renewcommand{\thefigure}{S\arabic{figure}}
\setcounter{table}{0}
\renewcommand{\thetable}{S\arabic{table}}
\section*{Supplementary figures and tables}

\begin{figure}[ht]
    \centering
    \includegraphics[width=\linewidth]{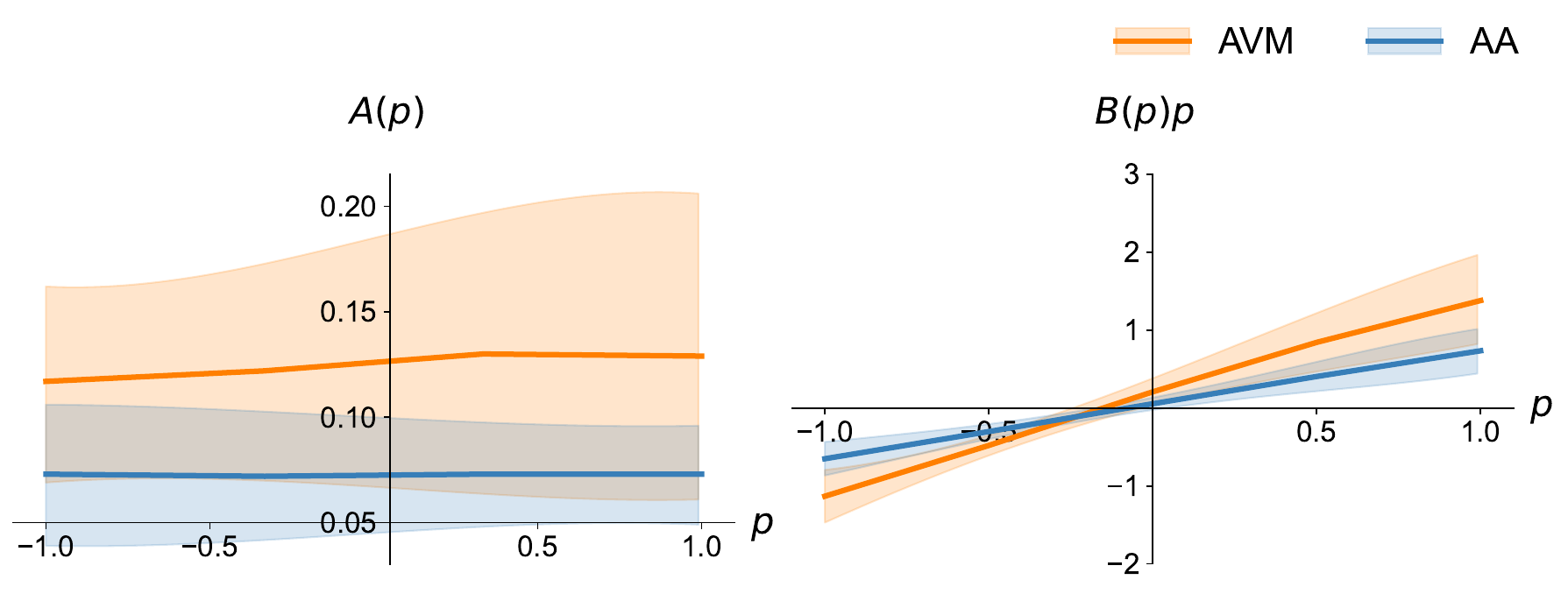}
    \caption{Shapes of polynomials $A(p)$ and $B(p) p$ with coefficients averaged over estimates made from the clinical data in Ref.~\citeonline{Cherevko2016}. The horizontal trend of $A(p) \approx a$ and linear trend of $B(p) p \approx b p$ suggest the modeling equation~\eqref{eq:model}. The error bands indicate standard deviations of the polynomial shapes.}
    \label{fig:si:polynomials}
\end{figure}

\begin{figure}[ht]
    \centering
    \includegraphics[width=\linewidth]{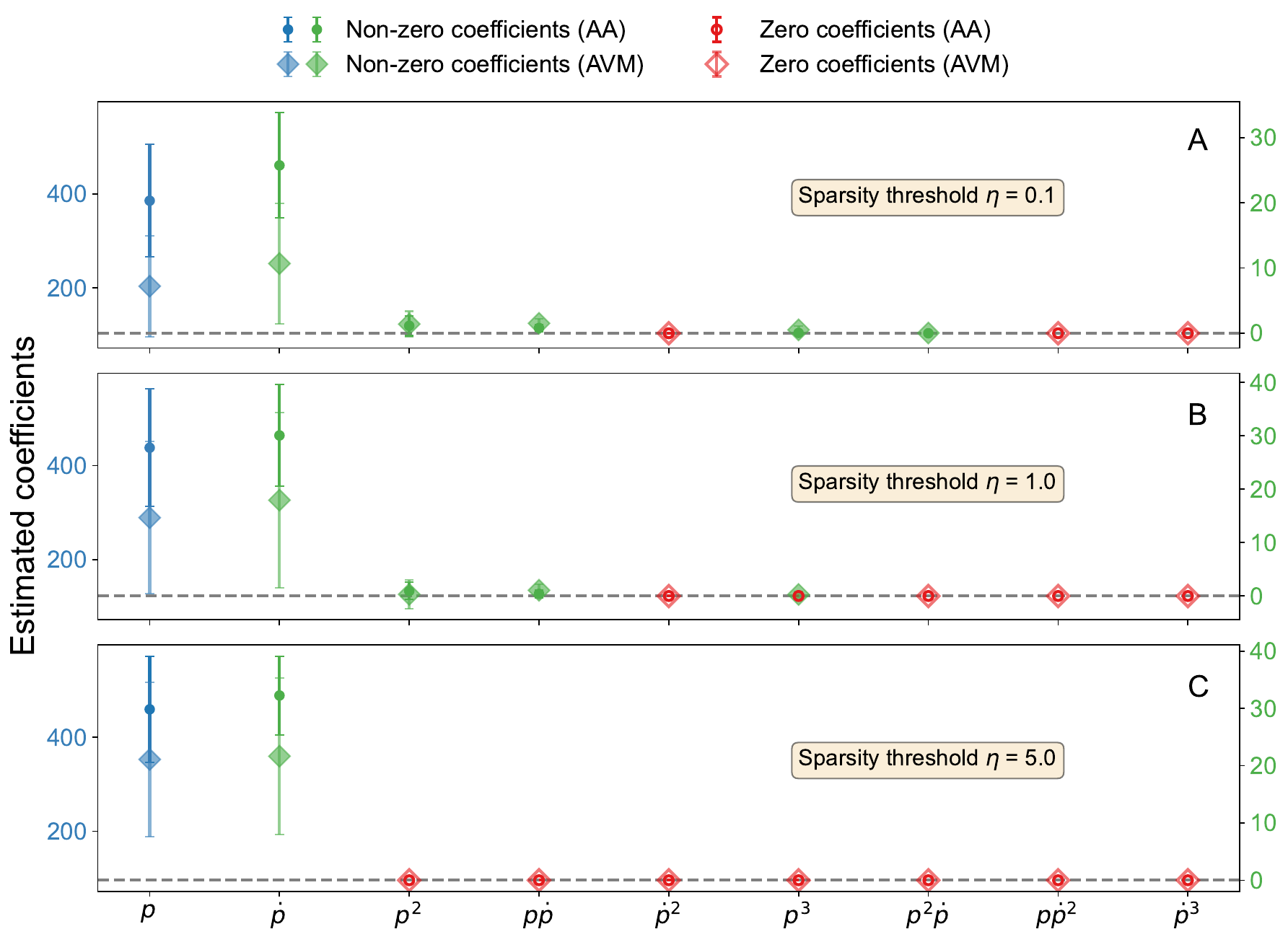}
    \caption{Estimated coefficients $c_{ij}$ of equation~\eqref{eq:general} averaged over all cases in the respective category of pathologies---AA (circles) or AVM (diamonds). Error bars are given by the standard deviations across the patients. The left vertical axis indicates scales of the coefficient $c_{10}$ (blue, \si{s^{-2}}), which is by order of magnitude larger than the other parameters with ticks on the right axis (green, various units). By increasing the sparsity threshold (A--C), we obtain sparser governing equations for pressure. The sparsity threshold $\eta = 5.0$ eliminates the coefficients of the higher-order polynomials in both malformations (red symbols), thus promoting the linear equation~\eqref{eq:model}.}
    \label{fig:si:fitting}
\end{figure}

\begin{figure}[ht]
    \centering
    \includegraphics[width=.7
    \linewidth]{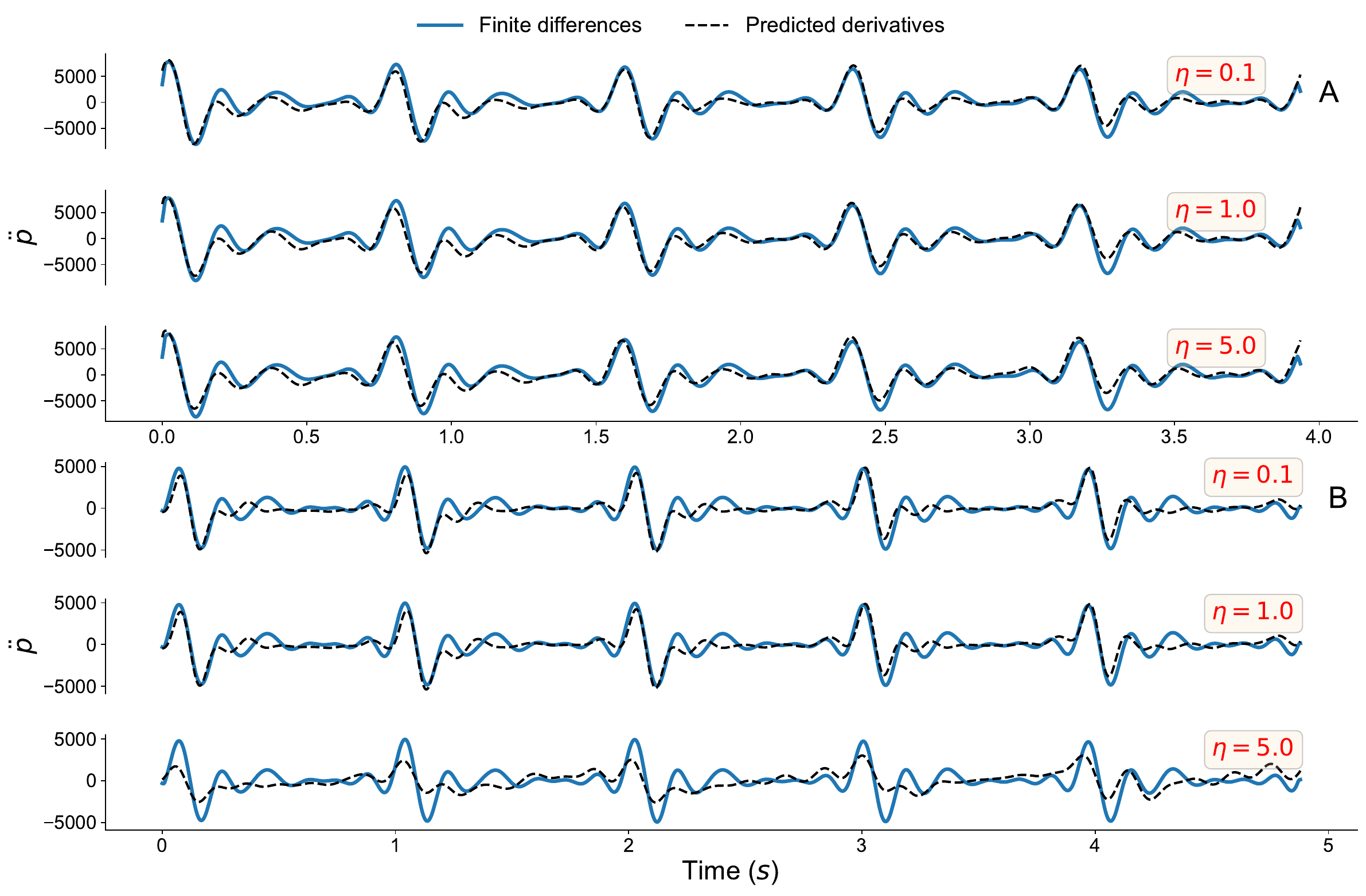}
    \caption{Comparison between the second time derivatives of pressure $\ddot{p}$ predicted by the best-fit models (dashed black lines) and the values obtained from finite differences of the experimental time series (solid blue lines). Our fitting procedure relies on the finite-difference estimates of $\ddot{p}$ [left-hand side of equation~\eqref{second-order}]. Predicted derivatives are calculated by evaluating the right-hand side of equation~\eqref{second-order} with the best-fit model parameters $\hat{\bm{\Xi}}$, the experimental values of $\left(p(t_i), v(t_i)\right)$, and the finite-difference estimate of $\dot{p}(t_i)$.
    For the same AA and AVM patients as in Figure~\ref{fig:library} (panels A and B respectively), higher-order models show better goodness of fit, even when they yield larger RMSEs for the simulated pressure.}
    \label{fig:si:derivatives}
\end{figure}

\begin{table}[ht]
\centering
\begin{tabular}{p{0.2\textwidth}p{0.2\textwidth}p{0.2\textwidth}}
\toprule
a ($\times$ \SI{e1}{Hz})  & b ($\times$ \SI{e2}{Hz^2}) & $\epsilon$ ($\times$ \SI{e4}{mmHg/ms}) \\
\midrule
\multicolumn{3}{c}{\textit{Before surgery}}\\
2.75 & 4.55 & 3.55 \\
3.66 & 5.08 & 4.96 \\
1.90 & 3.42 & 5.83 \\
3.68 & 6.41 & 9.44 \\
2.65 & 4.46 & 3.80 \\
\multicolumn{3}{c}{\textit{After surgery}}\\
3.59 & 5.08 & 10.11 \\
4.18 & 5.14 & 4.38 \\
3.51 & 4.59 & 4.80 \\
4.07 & 6.48 & 13.77 \\
2.81 & 3.09 & 3.04 \\
\bottomrule
\end{tabular}
\caption{\label{tab:1}Parameter values of equation~\eqref{eq:model} inferred from the AA patients' data}
\end{table}

\begin{table}[ht]
\centering
\begin{tabular}{p{0.2\textwidth}p{0.2\textwidth}p{0.2\textwidth}}
\toprule
a ($\times$ \SI{e1}{Hz})  & b ($\times$ \SI{e2}{Hz^2}) & $\epsilon$ ($\times$ \SI{e4}{mmHg/ms}) \\
\midrule
\multicolumn{3}{c}{\textit{Before surgery}}\\
 1.45 & 6.11 & 6.33 \\
 1.77 & 4.34 & 3.95 \\
 0.87 & 2.20 & 1.26 \\
 1.53 & 4.61 & 2.69 \\
 3.35 & 6.05 & 9.29 \\
\multicolumn{3}{c}{\textit{After surgery}}\\
 3.07 & 3.21 & 8.18 \\
 1.95 & 2.45 & 1.71 \\
 1.28 & 2.31 & 2.08 \\
 4.42 & 4.08 & 4.41 \\
 3.68 & 4.55 & 4.29 \\
\bottomrule
\end{tabular}
\caption{\label{tab:2}Parameter values of equation~\eqref{eq:model} inferred from the AVM patients' data}
\end{table}

\begin{figure}[ht]
    \centering
    \includegraphics[width=\linewidth]{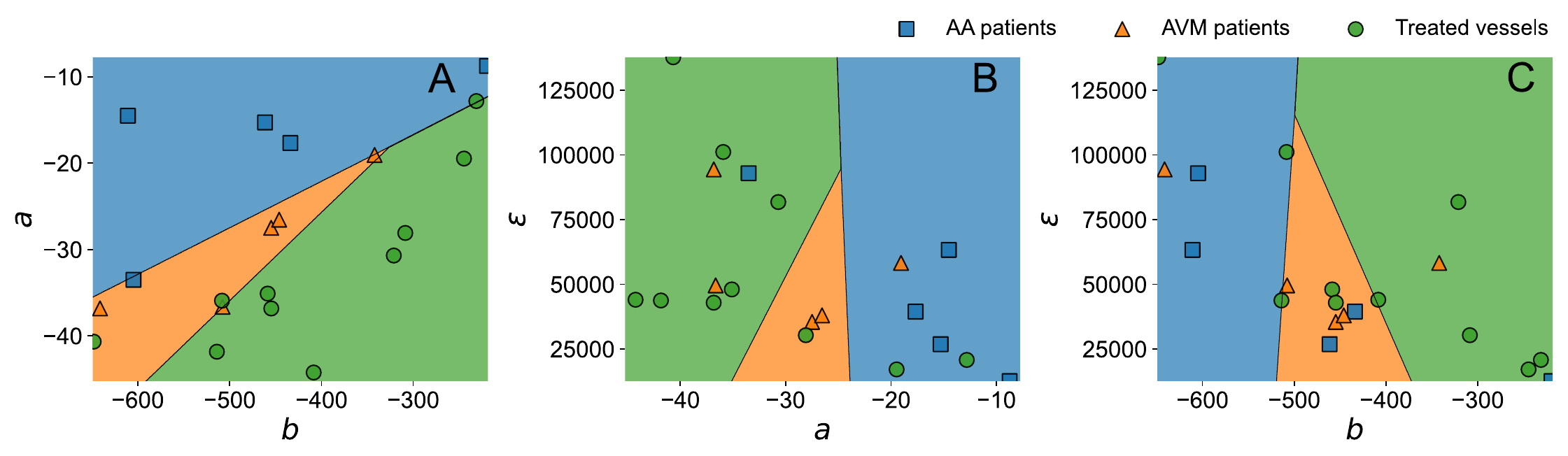}
    \caption{Decision boundaries of the logistic classifier equation~\eqref{eq:loglike} projected onto the $ab$-, $\epsilon{a}$-, and $\epsilon{b}$-planes cutting the third axis of the full $3$-dimensional feature space at the mean values of the parameters $\epsilon$, $b$, and $a$ over all patients' data respectively. The full dataset contains model parameters based on $5$ AA patients and $5$ AVM patients before surgery, the same patients' data after the surgery. The decision boundaries between these three classes are constructed by using MLxtend~\cite{raschkas_2018_mlxtend} python module.}
    \label{fig:si:classification2D}
\end{figure}

\begin{figure}[ht]
    \centering
    \includegraphics[width=.7\linewidth]{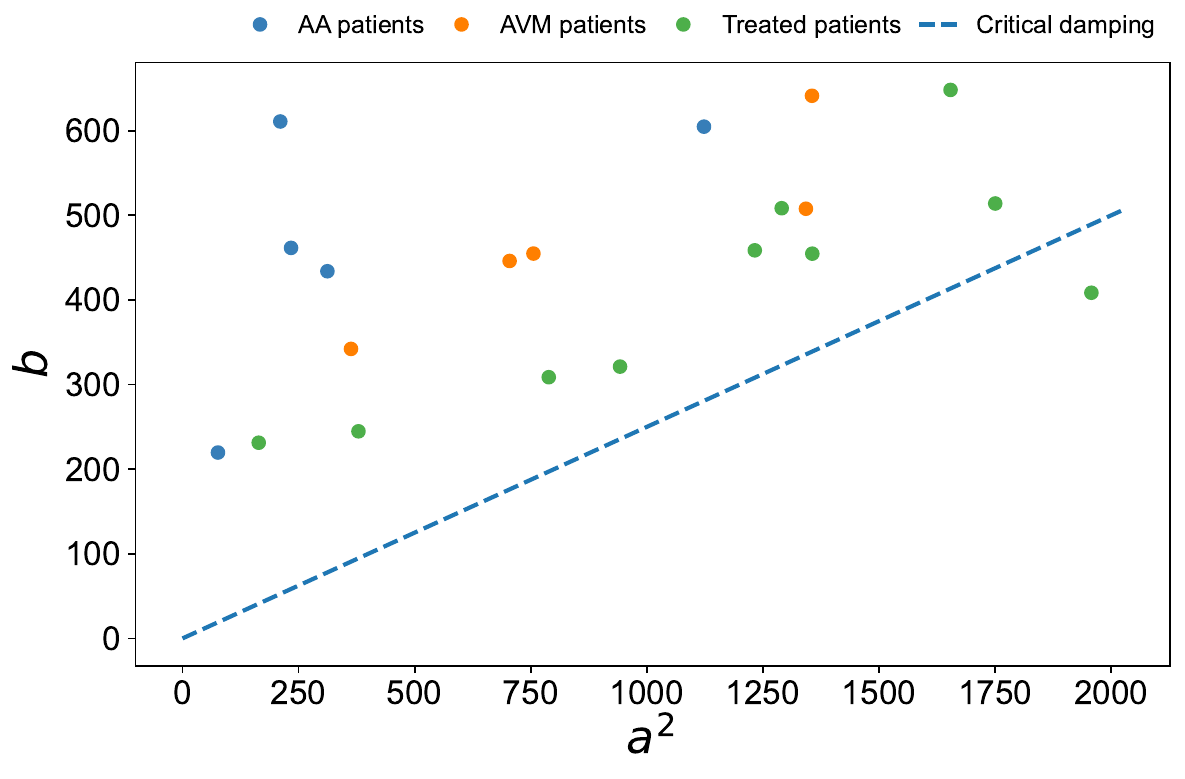}
    \caption{Parameters of the autonomous part (left-hand side) of equation~\eqref{eq:model} inferred from our clinical data. Most of the patients' data are described by the underdamped harmonic oscillator equation~\eqref{eq:model} with $b > a^2$. Treated patients tend to be closer to the line of critical damping $b = a^2$, whereas the AA patients tend to the region $b \gg a^2$, with AVM patients clustered between these two areas}
    \label{fig:si:damping}
\end{figure}

\end{document}